\documentclass[review]{elsarticle}

\usepackage[a4paper, total={6in, 9in}]{geometry}

\usepackage{hyperref}
\usepackage[utf8]{inputenc}
\usepackage{graphicx}
\graphicspath{{./figs/}}
\DeclareGraphicsExtensions{.pdf,.png}

\usepackage[ruled,lined,linesnumbered,commentsnumbered,longend]{algorithm2e}
\usepackage{amssymb}
\usepackage{amsmath}
\usepackage{multirow}
\usepackage{adjustbox}	
\usepackage{booktabs}
\usepackage{url}
\usepackage{hyperref}
\usepackage{float}
\usepackage[caption=false,font=footnotesize]{subfig}
\usepackage[table,xcdraw]{xcolor}

\DeclareMathOperator{\iSize}{ \textit{r} }
\DeclareMathOperator{\iEpochs}{ \textit{e}_{\textit{inc}} }

\journal{arXiv}

\begin{document}
\begin{frontmatter}

\title{Incremental Unsupervised Domain-Adversarial Training of Neural Networks}

\author[1]{Antonio-Javier Gallego\corref{corresponding}}
\cortext[corresponding]{Corresponding author}
\ead{jgallego@dlsi.ua.es}

\author[1]{Jorge Calvo-Zaragoza}
\ead{jcalvo@dlsi.ua.es}

\author[2]{Robert B. Fisher}
\ead{rbf@inf.ed.ac.uk}

\address[1]{Department of Software and Computing Systems, University of Alicante, \\03690 Alicante, Spain}

\address[2]{School of Informatics, University of Edinburgh, Edinburgh, EH8 9AB, UK}

\begin{abstract}
In the context of supervised statistical learning, it is typically assumed that the training set comes from the same distribution that draws the test samples. When this is not the case, the behavior of the learned model is unpredictable and becomes dependent upon the degree of similarity between the distribution of the training set and the distribution of the test set. One of the research topics that investigates this scenario is referred to as \emph{domain adaptation}. Deep neural networks brought dramatic advances in pattern recognition and that is why there have been many attempts to provide good domain adaptation algorithms for these models. Here we take a different avenue and approach the problem from an incremental point of view, where the model is adapted to the new domain iteratively. We make use of an existing unsupervised domain-adaptation algorithm to identify the target samples on which there is greater confidence about their true label. The output of the model is analyzed in different ways to determine the candidate samples. The selected set is then added to the source training set by considering the labels provided by the network as ground truth, and the process is repeated until all target samples are labelled. Our results report a clear improvement with respect to the non-incremental case in several datasets, also outperforming other state-of-the-art domain adaptation algorithms.
\end{abstract}

\begin{keyword}
	Domain Adaptation \sep
	Unsupervised learning \sep
	Neural Networks \sep
	Convolutional Neural Networks \sep
	Incremental labelling \sep
	Machine learning
\end{keyword}

\end{frontmatter}

\section{Introduction}
\label{sec:introduction}
Supervised learning is the most considered approach for dealing with classification tasks. This paradigm is based on a \emph{sufficiently representative} training set to learn a classification model. This level of representativeness is usually defined by two criteria: on the one hand, the training samples must be varied, which allows the algorithm to generalize instead of memorizing; on the other hand, the application of the trained model is assumed to be carried out on samples that come from the same distribution as those of the training set~\cite{DudaHS01}.

Building a training set fulfilling these conditions is not always straightforward. Although obtaining samples might be easy, assigning their correct labels is costly. This is why there are efforts to alleviate the aforementioned requirements. However, while the conflict between memorization and generalization has been well studied, and there exist established mechanisms to deal with it such as regularization or data augmentation~\cite{Goodfellow-et-al-2016}, learning a model that is able to correctly classify samples from a different target distribution remains open to further research. This problem is generally called \emph{transfer learning} (TL) \cite{shao2014transfer}, and when the classification labels do not vary in the target distribution it is usually referred to as \emph{domain adaptation} (DA) \cite{WANG2018135}.

Within the context of supervised learning, deep learning represents an important breakthrough \cite{lecun2015deep}. This term refers to the latest generation of artificial neural networks, for which novel mechanisms have been developed that allow training deeper networks, i.e., with many layers. These deep neural networks represent the state of the art in many classification tasks, and have managed to break the existing glass ceiling in many traditionally complex tasks. In turn, deep learning often requires a large amount of data, which makes the study of DA even more interesting.

As we will review in the next section, there are several alternatives to attempt DA, both general strategies and using deep neural networks. In this work we take a different avenue and study an incremental approach. We propose to use an existing DA algorithm for deep learning to classify those samples of the target domain for which the model is confident. Assuming the assigned labels as ground truth, the model is retrained. This added knowledge allows the network to refine its behavior to correctly classify other samples of the target set. This incremental process is repeated until the entire target set is completely annotated. We will show that this incremental approach achieves noticeable improvements with respect to the underlying DA algorithm. In addition, it is competitive on different benchmarks compared to other state-of-the-art DA algorithms.

The rest of the paper is structured as follows: we outline in Section 2 the existing literature about DA, with special emphasis to that based on deep neural networks; we present in Section 3 the proposed incremental methodology, as well as the underlying DA model that we consider in this work; we describe our experimental setting in Section 4, while the results are reported in Section 5; finally, the work is concluded in Section 6.

\section{Background}
\label{sec:background}
Since the beginning of machine learning research, there exists the idea of exploiting a model beyond its use over unknown samples of the source distribution. In the literature we can find two main topics that pursue this objective: the aforementioned TL and DA strategies.

In TL, some knowledge of the model is used to solve a different classification task. For example, a pre-trained DNN model can be used as initialization \cite{SimonyanZ14a,HeZRS16} or its feature extraction process can be considered as the basis of another classification model \cite{yosinski2014transferable}. As a special case of TL, the DA challenge typically assumes that the classification task of the target distribution is the same (i.e., the set of labels is equal). In this work we focus on the latter case.

In a DA scenario, we can also distinguish between semi-supervised and unsupervised approaches. While semi-supervised DA considers that some labeled samples of the target distribution are available \cite{6774431,yao2015semi,Saito2019}, unsupervised DA works with just unlabeled samples \cite{tpami_kouw}. We will revisit in this section unsupervised DA techniques, as it is the case of the proposed approach.

Performing unsupervised DA is still considered an open problem from both theoretical and practical perspectives \cite{BousmalisSDEK17}. Most approaches consider that the key is to build a good feature representation that becomes invariant to the domain \cite{Ben-David2006}. A good example is the \emph{Domain Adaptation Neural Network} (DANN) proposed by Ganin et al. \cite{Ganin2016DANN}, which simultaneously learns domain-invariant features from both source and target data and discriminative features from the source domain. Following this line of research, many approaches have been proposed more recently: \emph{Virtual Adversarial Domain Adaptation} (VADA) proposed by Shu et al. \cite{shu2018a} added a penalty term to the loss function to penalize class boundaries that cross high-density feature regions. The \emph{Deep Reconstruction-Classification Networks} (DRCN) \cite{Ghifary2016} consists of a neural network that forces a common representation of both the source and target domains by sample reconstruction, while learning the classification task from the source samples. The \emph{Domain Separation Networks} (DSN) proposed by Bousmalis et al. \cite{Bousmalis2016} are trained to map input representations onto both a domain-specific subspace and a domain-independent subspace, in order to improve the way that the domain-invariant features are learned. Haeusser et al. \cite{Haeusser2017} proposed \emph{Associative Domain Adaptation} (ADA), which is another domain-invariant feature learning approach that reinforces associations between source and target representations in an embedding space with neural networks. The \emph{Adversarial Discriminative Domain Adaptation} (ADDA) strategy \cite{Tzeng2017} follows the idea of Generative Adversarial Networks, along with discriminative modeling and untied weight sharing to learn domain-invariant features, while keeping a useful representation for the discriminative task. \emph{Drop to Adapt} (DTA) \cite{Lee2019} makes use of adversarial dropout to enforce discriminative domain-invariant features. Damodaran et al. \cite{Damodaran2018} proposed the \emph{Deep Joint Distribution Optimal Transport} (DeepJDOT) approach, which learns both the classifier and aligned data representations between the source and target domain following a single neural framework with a loss functions based on the Optimal Transport theory \cite{optimal-transport}.

A different strategy to DA consists in learning how to transform features from one domain to another. Following this idea, the \emph{Subspace Alignment} (SA) method \cite{Fernando2013SA} seeks to represent the source and target domains using subspaces modelled by eigenvectors. Then, it solves an optimization problem to align the source subspace with the target one. Also, Sun and Saenko proposed the \emph{Deep Correlation Alignment} (D-CORAL) approach \cite{Baochen2016CORAL}, which consists of a neural network that learns a nonlinear transformation to align correlations of layer activations from the source and target distributions.

While the methods outlined above seek for new ways to achieve the desired characteristics of a proper DA method, our proposed approach takes a different avenue. Specifically, we build upon the existing DANN approach, and we propose novel ways to improve its ability to adapt to the target domain by performing the adaptation incrementally.



%
%
%
%
%
%
%

\section{Methodology}
\label{sec:method}

\subsection{Preliminaries}
Let $X$ be the input space and $Y$ be the output or \emph{label} space. A classification task assumes that there exist a function $f : X \rightarrow Y$ that assigns a label to each possible sample of the input space. For supervised learning, the goal is to learn a hypothesis function $h$ that models the unknown function $f$ with the least possible error. We refer to $h$ as label classifier. Quite often, the approach is to estimate a posterior probability $P(Y|X)$ so that the label classifier follows a \emph{maximum a posteriori} decision such that $h(x) = \arg\max_{y \in Y} P(x|y)$. This is the case with neural networks.

In the DA scenario, there exist two distributions over $X \times Y$: $D_S$ and $D_T$, which are referred to as \emph{source domain} and \emph{target domain}, respectively. We focus on the case of unsupervised domain adaptation, for which DA is only provided with a labeled source set $S=\left \lbrace \left ( x_i,y_i\right )\right \rbrace _{i=1}^n\thicksim (D_S)^n$ and a completely unlabeled target domain $T=\left \lbrace \left (x_i\right )\right \rbrace _{i=1}^{n'}\thicksim (D_T)^{n'}$.

The goal of a DA algorithm is to build a label classifier for $D_{T}$ by using the information provided in both $S$ and $T$.

\subsection{Domain Adaptation Neural Network}
Given its importance in the context of our work, we further describe here the operation of DANN, which will be considered as the backbone for our incremental approach.

DANN is based on the \emph{theory of learning from different domains} discussed by \cite{BenDavid2006,BenDavid2010}. This suggests that the transfer of the knowledge gained from one domain to another must be based on learning features that do not allow to discriminate between the two domains (source and target) of the samples to be classified. For this, DANN learns a classification model from features that do not encode information about the domain of the sample to be classified, thus generalizing the knowledge from a source labeled domain to a target unlabeled domain.

More specifically, the proposed neural architecture includes a \emph{feature extractor} module ($G_f$) and a \emph{label classifier} ($G_y$), which together build a standard feed-forward neural network that can be trained to classify an input sample $x$ into one of the possible categories of the output space $Y$. The last layer of the label classifier $G_y$ uses a ``softmax'' activation, which models the posterior probability $P(x \mid y),\, \forall y \in Y$ of a given input $x \in X$.

DANN adds a new \emph{domain classifier} module ($G_d$) to the neural network, that classifies the domain to which the input sample $x$ belongs. This classifier is built as a binary logistic regressor that models the probability that an input sample $x$ comes from the source distribution ($d_i = 0$ if $x \sim{\cal D}_S$) or the target distribution ($d_i = 1$ if $x \sim{\cal D}_T$), where $d_i$ denotes a binary variable that indicates the domain of the sample.

The unsupervised adaptation to a target domain is achieved as follows: the domain classifier $G_d$ is connected to the feature extractor $G_f$ (which is shared with the label classifier $G_y$) through the so-called \emph{gradient reversal layer} (GRL). This layer does nothing at prediction. However, while learning through back-propagation, it multiplies the gradient by a certain negative constant ($\lambda$). In other words, the GRL receives the gradient from the subsequent layer and multiplies it by $-\lambda$, therefore changing its sign before passing it to the preceding layer.
The idea of this operation is to force $G_f$ to learn generic features that do not allow discriminating the domain.
In addition, since this training is carried out simultaneously with the training of $G_y$ (label classifier), the features must be adequate for discriminating the categories to classify, yet unbiased with respect to the input domain. According to the DA theory, this should cause $G_y$ to be able to correctly classify input samples regardless of their domain, given that the features from $G_f$ are forced to be invariant.

The DANN training simultaneously updates all modules, providing samples for both $G_y$ and $G_d$. This can be done by using conventional mechanisms such as Stochastic Gradient Descent, from batches that include half of the examples from each domain. During the training process, the learning of $G_f$ pursues a trade-off between appropriate features for the classification ($G_y$) and inappropriate features for discriminating the domain of the input sample ($G_d$). The hyper-parameter $\lambda$ allows tuning this trade-off. The training is performed until the result converges to a saddle point, which can be found as a stationary point in the gradient update defined by the following equation:

\begin{equation}
    \theta_f \leftarrow \theta_f
        - \mu \left(
            \frac{\partial {\cal L}_y}{\partial \theta_f}
            - \lambda
            \frac{\partial {\cal L}_d}{\partial \theta_f}
        \right)
\label{eq:dann_loss}
\end{equation}

\noindent where $\theta_f$ denotes the weights of $G_f$, $\mu$ denotes the learning rate, and ${\cal L}_y$ and ${\cal L}_d$ represent the loss functions for the label classifier and the domain classifier, respectively.

\begin{figure*}[ht]
  \centering
  \includegraphics[width=.8\linewidth]{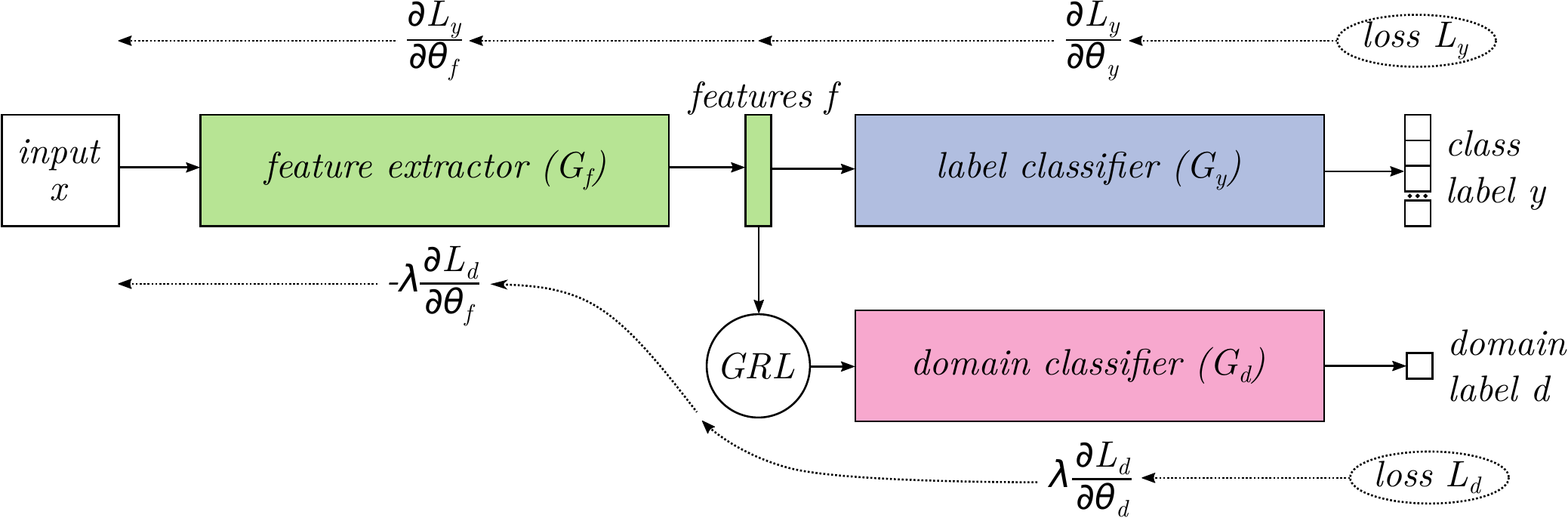}
  \caption{Graphical overview of the DANN architecture, consisting of three blocks: feature extractor ($G_f$), label classifier ($G_y$), and domain classifier ($G_d$). The GRL circle denotes the gradient reversal layer that multiplies the gradient by a negative factor.}
  \label{fig:dann_scheme}
\end{figure*}

A graphical overview of the DANN architecture is depicted in Fig. \ref{fig:dann_scheme}.

\subsection{Incremental DANN}
Our main contribution within the context of DA is to propose an incremental approach to DANN (iDANN). This strategy is explained below.

Once the DANN model is trained as explained in the previous section, we can use both the feature extractor $G_f$ and the label classifier $G_y$ to predict the category of samples from both the target domain and the source domain ($G_y(G_f(x))$). The ``softmax'' activation used at the output of this classifier returns the posterior probability that the network considers $x$ to belong to any of the classes of the output space $Y$.

Our main assumption is that we can use the subset of samples from the target domain for which $G_y$ is more confident, and then add them to the source labeled domain assuming the prediction as ground truth. These samples are thereafter considered as samples of the source domain completely. Afterwards, we can retrain the DANN network to fine-tune its weights using the new training set. This process is repeated iteratively, moving the labeled samples with greater confidence from the target domain to the source domain after each iteration. We stop when there are no more samples to move from the target domain.

The intuitive idea behind our approach is that by adding target domain information to the source (labeled) domain, the DANN learns new domain-invariant features that better fit the eventual classification task, thereby becoming more accurate for other target domain samples. In each iteration, however, the task increases its complexity because it deals first with the simplest samples to classify (for which the DANN is more confident), leaving those that have more dissimilar features in the unlabeled target set.
When the DANN is retrained with labeled samples that include target domain information, the domain classifier $G_d$ needs to be more specific. This forces the feature extraction module $G_f$ to forget the features that differentiate more complex samples from the target domain.

We formalize the process in Algorithm \ref{alg:idann}, where
$e$ and $b$ represents the number of epochs and the batch size considered, respectively,
$\iEpochs$ denotes the number of epochs for the incremental stage of the algorithm,
$\iSize$ indicates the size of the subset of target domain samples to select in each iteration,
and $\beta$ is a constant that allows us to modify this size after each iteration.

\begin{algorithm}[ht]
\caption{Incremental DANN (iDANN)}
\label{alg:idann}
\SetAlgoLined
\SetKwInOut{KwIn}{Input}
\SetKwInOut{KwOut}{Output}
\KwIn{
    $S \leftarrow \{( x_i,y_i) \sim {\cal D}_S \}$ \\
    $T \leftarrow \{(x_i) \sim {\cal D}_T \}$ \\
    $e, \iEpochs, b, \iSize, \lambda, \beta \leftarrow
    \text{Initial hyper-parameters values}$ \\
}
\KwOut{$G_f, G_y, \text{CNN}$}
\While{$T \neq \emptyset$}{
    $G_f, G_y \leftarrow \text{Fit DANN with } \{S, T, e, b, \lambda\}$ \\
    $\hat{B}_{\iSize} \leftarrow \text{\texttt{selection\_policy}} (G_f, G_y, T, \iSize)$ \\
    $S \leftarrow S \cup \hat{B}_{\iSize}$ \\
    $T \leftarrow T \backslash \hat{B}_{\iSize}$ \\
    $e \leftarrow \iEpochs$ \\
    $\iSize \leftarrow \beta \iSize$ \\
}
$\hat{T} \leftarrow \{ ( x_i,y_i) \, \mid \, x_i \sim {\cal D}_T, y_i = G_y( G_f( x_i ) ) \}  $ \\
$\text{Fit CNN with } \{ \hat{T}, e, b \}$ \\
\end{algorithm}


In this algorithm, the samples of the target domain ($\hat{B}$) are classified using the label classifier $G_y$, and then it proceeds to select a subset $\hat{B}_{\iSize}$ of size $\iSize$ to be moved from the target domain to the source domain. For this purpose, two selection criteria are proposed, which are described in the next section.

Once the iterative stage of the algorithm ends, the label classifier $G_y$ is used to classify the entire original target domain (see line 9 of Algorithm \ref{alg:idann}). This labeled target set is used to then train a neural network from scratch, which is therefore specialized in classifying target domain samples (more details in Section \ref{sec:separated_cnn}).

\subsection{Selection policies}
Below we describe in detail the two proposed policies to select samples during the iterative stage of Algorithm~\ref{alg:idann} (\texttt{selection\_policy}). One policy is directly based on the confidence level that the network provides to the prediction, while the other is based on geometric properties of the learned feature space.

\subsubsection{Confidence policy
\label{sec:prob_policy}}


As mentioned above, the output of the label classifier $G_y$ uses a softmax activation. Let $L$ denote the number of labels. Then, the standard softmax function $\sigma : \mathbb{R}^L \to \mathbb{R}^L$ is defined by Equation \ref{eq:softmax}.

\begin{equation}
\sigma(\mathbf{z})_i = \frac{e^{z_i}}{\sum_{j=1}^L e^{z_j}}
\text{ for } i = 1, \dotsc, L
\text{ and } \mathbf{z} = (z_1, \dotsc, z_L) \in \mathbb{R}^L
\label{eq:softmax}
\end{equation}

This function normalizes an $L$-dimensional vector $\mathbf{z}$ of unbounded real values into another $L$-dimensional vector $\sigma(\mathbf{z})$, for which values range between $[0,1]$ and add up to $1$. This can be interpreted as a posterior probability over the different possible labels \cite{Bridle1990softmax}. In order to turn these probabilities into the predicted class label, we simply take the argmax-index position of this output vector, following a \emph{Maximum a Posteriori} probability criterion.

Taking advantage of this interpretation, the first policy for selecting samples to move from the target domain to the source is based on the probability provided by the label classifier $G_y$, which can be seen as a measure of confidence in such classification.

With this criterion, we will keep the maximum predicted probability value for each sample of the target set among the possible labels. Then, we will order all samples based on this value---from highest to lowest---in order to select the first $\iSize$ samples to build the subset $\hat{B}_{\iSize}$.

Algorithm \ref{alg:prob_policy} presents the algorithmic description of this process, where $G^{p}_y$ refers to the probabilistic output of the label classifier after the softmax activation, before applying argmax to select a label. The function \textit{sortr} is used to sort the set in decreasing order.

\begin{algorithm}[ht]
\caption{Confidence policy}
\label{alg:prob_policy}
\SetAlgoLined
\SetKwInOut{KwIn}{Input}
\SetKwInOut{KwOut}{Output}
\KwIn{
    $T \leftarrow \{(x_i) \sim {\cal D}_T \}$ \\
    $G_f, G_y \leftarrow \text{Feature extractor and label classifier}$ \\
    $\iSize \leftarrow \text{Size of the selected samples subset}$ \\
}
\KwOut{$\hat{B}_{\iSize}$}
$\hat{B} \leftarrow G^{p}_y( G_f( T ) )$ \\
$\hat{B}_{\iSize} \leftarrow \{\emptyset\}$ \\
\ForEach{$(x_i,y_i) \in \textit{sortr}( \hat{B} )$}{
  $\hat{B}_{\iSize} \leftarrow \hat{B}_{\iSize} \cup (x_i,y_i)$ \\
  \If{$|\hat{B}_{\iSize}| = \iSize$}{
    break \\
  }
}
\end{algorithm}

\begin{figure*}[ht]
  \centering
  \includegraphics[width=.48\linewidth]{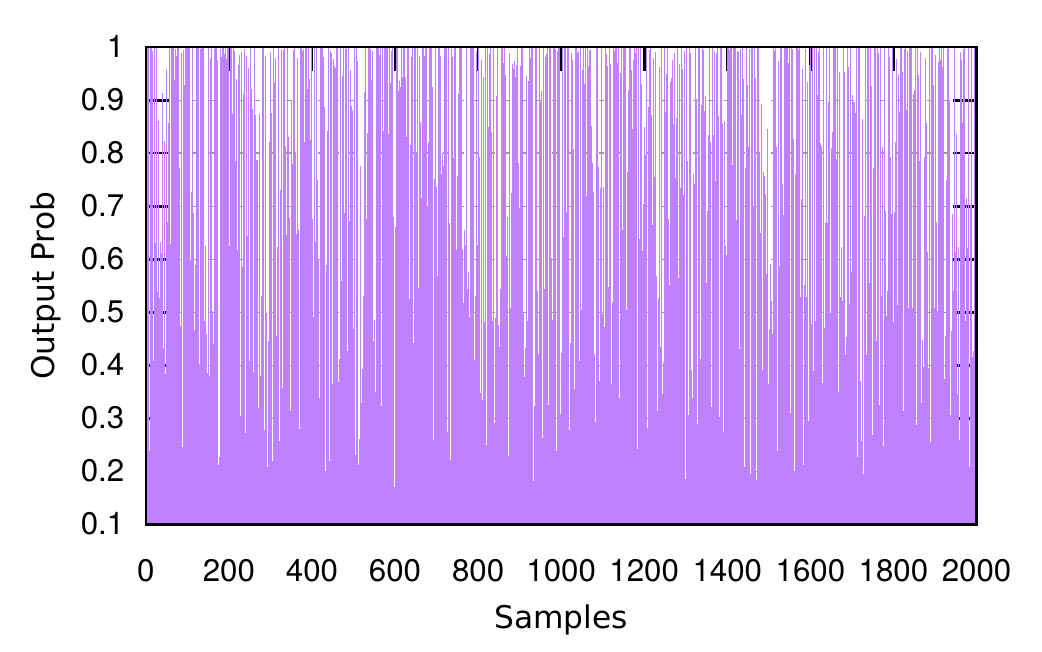}
  \includegraphics[width=.48\linewidth]{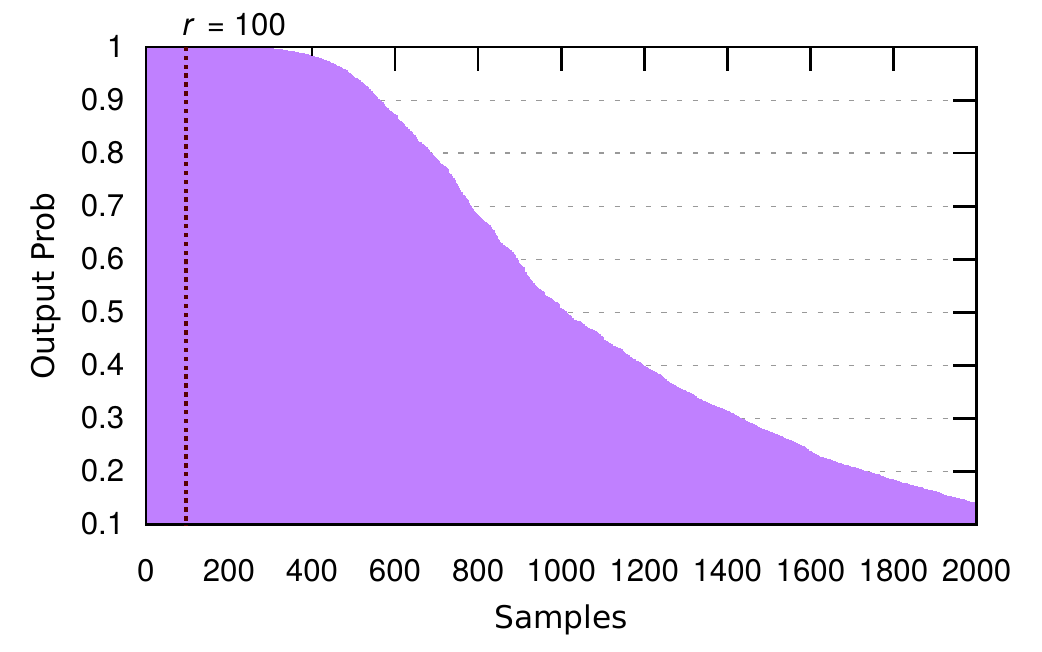}
  \caption{Example of probabilities obtained with DANN. Left: maximum probability of each sample. Right: ordered set of maximum probabilities, where the threshold $\iSize$ has been highlighted.
  }
  \label{fig:policy1_prob}
\end{figure*}

Figure \ref{fig:policy1_prob} shows an example of a set of probabilities obtained after predicting the target samples with DANN. The figure on the left shows the maximum probability values obtained for the classification of each sample---without sorting---while in the figure on the right the sorted set is shown, where the threshold $\iSize$ has been highlighted.

\subsubsection{$k$NN policy
\label{sec:knn_policy}}

As in the previous case, once the network has been trained, we use the label classifier $G_y$ to predict the labels of the whole target domain and then we sort them based on the confidence given by the network. However, in this case, instead of directly selecting a subset of samples according to this confidence, we will also evaluate the geometric properties of the feature space. This is performed following the $k$-nearest neighbor rule.

We first obtain the feature set $F_S$ from the source set $S$ (using $G_f(S)$). We then proceed to iterate the target set samples sorted by their level of confidence. Given a target sample, if the label of the $k$-nearest samples of the source domain matches the label assigned by the label classifier $G_y$, then we will select the prototype. Otherwise, we will discard it. Therefore, samples are selected based on both the confidence provided by the DANN in their label and the extent they match the distribution of the source domain.

Algorithm \ref{alg:knn_policy} describes this process algorithmically. The $kNN (q, F_S, k) $ function receives as parameters the query sample $q$, the set $F_S$ and the value $k$ to be used, and yields the predicted label $l$ and the number of samples $m$ within its $k$-nearest neighbors from $S$ that have the same label.

\begin{algorithm}[ht]
\caption{$k$NN policy}
\label{alg:knn_policy}
\SetAlgoLined
\SetKwInOut{KwIn}{Input}
\SetKwInOut{KwOut}{Output}
\KwIn{
    $S \leftarrow \{( x_i,y_i) \sim {\cal D}_S \}$ \\
    $T \leftarrow \{(x_i) \sim {\cal D}_T \}$ \\
    $G_f, G_y \leftarrow \text{Feature extractor and label classifier}$ \\
    $\iSize \leftarrow \text{Size of the selected samples subset}$ \\
    $k \leftarrow \text{Number of neighbors to consider}$ \\
}
\KwOut{$\hat{B}_{\iSize}$}
$F_S \leftarrow G_f( S )$ \\
$\hat{B} \leftarrow G^{p}_y( G_f( T ) )$ \\
$\hat{B}_{\iSize} \leftarrow \{\emptyset\}$ \\
\ForEach{$(x_i,y_i) \in \textit{sortr}( \hat{B} )$}{
  $f^{(i)}_{T} \leftarrow G_f( x_i )$ \\
  $l, m \leftarrow kNN(f^{(i)}_{T}, F_S, k)$ \\
  \If{$y_i = l \textbf{ and } m = k$}{
    $\hat{B}_{\iSize} \leftarrow \hat{B}_{\iSize} \cup (x_i,y_i)$ \\
    \If{$|\hat{B}_{\iSize}| = \iSize$}{
        break \\
    }
  }
}
\end{algorithm}

\begin{figure*}[ht]
  \centering
  \includegraphics[width=.5\linewidth]{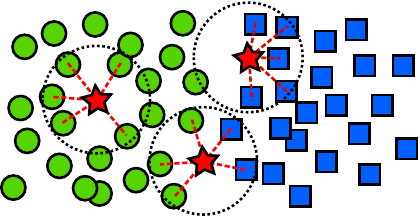}
  \caption{Example of sample selection using the kNN policy with $k = 5$. Green circles and blue squares represent samples of two different classes from the source domain. Red stars represent the samples of the target domain that are evaluated to determine whether they are chosen.}
  \label{fig:knn_policy}
\end{figure*}

The idea of this policy is to select the samples of the target domain whose features are within the cluster of the source domain for the same class. An illustrative example of this condition is shown in Fig.~\ref{fig:knn_policy} with $k = 5$. The example shows two labels of the source domain as green circles and blue squares. The red stars denote the target domain examples that are being evaluated to determine if they are selected. For instance, the star on the left would be selected if, and only if, the network classified it as a green circle, since its 5-nearest neighbors are green circles. Similarly, the star on the right would be selected if, and only if, the network classified it as a blue square. However, the central star would always be discarded because its 5-neighbors belong to two different classes.

If we increase $k$, the red star of the left would still be selected (if labeled as green circle) because it is located in the middle of the cluster. However, the red start of the right is closer to label boundaries, and so it would eventually be discarded.

\subsection{Training a CNN with the new labeled target set
\label{sec:separated_cnn}}
As described in Algorithm \ref{alg:idann}, once the iterative stage of the iDANN algorithm is completed, we use the label classifier $G_y$ to annotate the entire original target set $T$ from scratch. Then, a new CNN is trained by conventional means considering the same neural architecture of $G_y(G_f(\cdot))$. This allows us to eventually get a neural network that is directly specialized in the classification of the target domain.

However, we assume that some part of the iterative annotation of the target set will contain noise at the label level. To mitigate the possible efects of this noise, we consider \emph{label smoothing} \cite{Szegedy2016}. This is an efficient and theoretically-grounded strategy for dealing with label noise, which also makes the model less prone to overfitting.

Compared to classical one-hot output representation, label smoothing changes the construction of the true probability to

\begin{equation}
y'_i = (1 - \epsilon) y_i + \frac{\epsilon}{L},
\label{eq:label_smooth}
\end{equation}

where $\epsilon$ is a small constant (or smoothing parameter) and $L$ is the total number of classes. Hence, instead of minimizing cross-entropy with hard targets (0 or 1), it considers soft targets.








\section{Experimental setup}
\label{sec:setup}

\subsection{Datasets}

The proposed approach will be evaluated with two different classification tasks, that are common in the DA literature. The first one is that of digit classification, for which we consider the following datasets:

\begin{itemize}
    \item MNIST \cite{LeCun1998mnist}: this collection contains $28\times 28$ images representing isolated handwritten digits.
    \item MNIST-M \cite{Ganin2016DANN}: this dataset was synthetically generated by merging MNIST samples with random color patches from BSDS500~\cite{Arbelaez2011BSDS500}.
    %
    \item Street View House Numbers (SVHN) \cite{Netzer2011SVHN}: it consists of images obtained from house numbers from Google Street View. It represents a real-world challenge of digit recognition in natural scenes, for which several digits might appear in the same image and only the central one must be classified.

    \item Synthetic Numbers \cite{Ganin2016DANN}: images of digits generated using Windows{\texttrademark} fonts, with varying position, orientation, color and resolution.
\end{itemize}

In addition, we also evaluate our approach for traffic sign classification with the following datasets:

\begin{itemize}
    \item German Traffic Sign Recognition Benchmark (GTSRB) \cite{Stallkamp2012GTSRB}: this dataset contains images of traffic signs obtained from the real world in different sizes, positions, and lighting conditions, as well as including occlusions.
    \item Synthetic Signs \cite{Moiseev2013SynthSigns}: this dataset was synthetically generated by taking common street signs from Wikipedia and applying several transformations. It tries to simulate images from GTSRB although there are significant differences between them.
\end{itemize}

Table \ref{tab:datasets} summarizes the information of our evaluation corpora, including the domain to which they belong, the number of labels, the image resolution, the number of samples, and the type of image indicating whether they are in color or grayscale format. Figure \ref{fig:datasets} shows some random examples from each of these datasets.

\begin{table}[ht]
\caption{Description of the datasets used in the experimentation.}
\label{tab:datasets}
\centering
\begin{adjustbox}{width=\textwidth, keepaspectratio}
\begin{tabular}{ccccccc}
\toprule
    \textbf{Set} &
    \textbf{\# labels} &
    \textbf{Domain} &
    \textbf{Resolution (px)} &
    \textbf{Gray/Color} &
    \textbf{\# samples} & \\
\hline
\multirow{4}{*}{Numbers}
 & \multirow{4}{*}{10}
   & MNIST        & $28\times28$  & Gray  & 65,000 \\
 & & MNIST-M      & $28\times28$  & Color & 65,000 \\
 & & SVHN         & $32\times32$  & Color & 99,289 \\
 & & Syn. Numbers & $32\times32$  & Color & 488,953 \\
\hline
\multirow{2}{*}{\begin{tabular}[c]{@{}c@{}}Traffic\\signs\end{tabular}}
 & \multirow{2}{*}{43}
   & GTSRB        & $[25\times25, 225\times243]$   & Color  & 51,839 \\
 & & Syn. Signs   & $40\times40$   & Color  & 100,000 \\
\bottomrule
\end{tabular}
\end{adjustbox}
\end{table}

\newcommand{\figDatasetsSize}{.2}
\newcommand{\figDatasetsHspace}{0.8cm}
\begin{figure}[ht]
\centering
  \subfloat[MNIST]{
  	\includegraphics[width=\figDatasetsSize\linewidth]{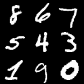}
     \label{fig:dataset_mnist}
  }
  \hspace{\figDatasetsHspace}
  \subfloat[MNIST-M]{
  	\includegraphics[width=\figDatasetsSize\linewidth]{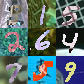}
     \label{fig:dataset_mnist_m}
  }
  \hspace{\figDatasetsHspace}
  \subfloat[SVHN]{
  	\includegraphics[width=\figDatasetsSize\linewidth]{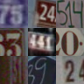}
     \label{fig:dataset_svhn}
  }
  \\
  \subfloat[Syn. Numbers]{
  	\includegraphics[width=\figDatasetsSize\linewidth]{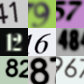}
     \label{fig:dataset_syn_numbers}
  }
  \hspace{\figDatasetsHspace}
  \subfloat[GTSRB]{
  	\includegraphics[width=\figDatasetsSize\linewidth]{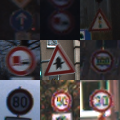}
     \label{fig:dataset_gtsrb}
  }
  \hspace{\figDatasetsHspace}
  \subfloat[Syn. Signs]{
  	\includegraphics[width=\figDatasetsSize\linewidth]{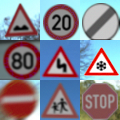}
     \label{fig:dataset_syn_signs}
  }
\caption{Random examples from the datasets used in experimentation.}
\label{fig:datasets}
\end{figure}

The images of each classification task were rescaled to the same size: the digits to $28 \times 28$ pixels, and the traffic signs to $40 \times 40$ pixels. Concerning the pre-processing of the input data, the images were normalized within the range $[0,1]$.  The train and test partitions were those proposed by the authors of each dataset, in order to ensure a fair comparison with the results obtained in the literature.

\subsection{CNN architectures}
To evaluate the proposed methodology, the same three CNN architectures used in the original DANN paper have been tested. Table \ref{tab:topologies} reports a summary of these architectures.

As the authors pointed out, these topologies are not necessarily optimal and better adaptation performance might be attained if they were tweaked. However, we chose to keep the same configuration to make a fairer comparison.

As the activation function, a Rectifier Linear Unit (ReLU) was used for each convolution layer and fully-connected layer, except for the the output layers. $L$ neurons with softmax activation were used as output of the label classifier. For the output of the domain classifier, a single neuron with a logistic (sigmoid) activation function was used to discriminate between two possible categories (source domain or target domain).

Model 1 was used for all the experiments with digit datasets, except those using SVHN. This topology is inspired by the classical LeNet-5 architecture \cite{LeCun1998mnist}. Model 2 was used to evaluate the experiments with digits that include SVHN. This architecture is inspired by \cite{Srivastava2014}. Finally, Model 3 was used for the experiments with traffic sings. In this case, the single-CNN baseline obtained from \cite{Ciresan2012} was used.

\begin{table}[ht]
\caption{CNN network configurations considered. Notation: $Conv(f,w,h)$ stands for a layer with $f$ convolution operators with a kernel of size $w\times h$ pixels, $MaxPool(w,h)$ stands for the max-pooling operator of dimensions $w\times h$ pixels---with a $2\times2$ in all cases---and $FC(n)$ represents a fully-connected layer of $n$ neurons. In the output layer of the label classifier a fully-connected layer of $L$ neurons with softmax activation is added, where $L$ denotes the number of categories of the dataset at issue.}
\label{tab:topologies}
\centering
\begin{adjustbox}{width=\textwidth, keepaspectratio}
\begin{tabular}{c|lll|l|l}
\toprule
 & \multicolumn{3}{c|}{\textbf{Feature extractor}} & & \\ 
\textbf{Model}
    & \multicolumn{1}{c}{\textbf{Layer 1}}
    & \multicolumn{1}{c}{\textbf{Layer 2}}
    & \multicolumn{1}{c|}{\textbf{Layer 3}}
    & \multicolumn{1}{c|}{\textbf{Label classifier}}
    & \multicolumn{1}{c}{\textbf{Domain classifier}}
\\ \hline
1
& \begin{tabular}[c]{@{}l@{}}Conv(32, 5, 5)\\MaxPool(2, 2)\end{tabular}
& \begin{tabular}[c]{@{}l@{}}Conv(48, 5, 5)\\MaxPool(2, 2)\end{tabular}
&
& \begin{tabular}[c]{@{}l@{}}FC(100)\\FC(100)\\FC(L)\end{tabular}
& \begin{tabular}[c]{@{}l@{}}FC(100)\\FC(1)\end{tabular}
\\ \hline
2
& \begin{tabular}[c]{@{}l@{}}Conv(64, 5, 5)\\MaxPool(3, 3)\end{tabular}
& \begin{tabular}[c]{@{}l@{}}Conv(64, 5, 5)\\MaxPool(3, 3)\end{tabular}
& \begin{tabular}[c]{@{}l@{}}Conv(128, 5, 5)\end{tabular}
& \begin{tabular}[c]{@{}l@{}}FC(3072)\\FC(2048)\\FC(L)\end{tabular}
& \begin{tabular}[c]{@{}l@{}}FC(1024)\\FC(1024)\\FC(1)\end{tabular}
\\ \hline
3
& \begin{tabular}[c]{@{}l@{}}Conv(96, 5, 5)\\MaxPool(2, 2)\end{tabular}
& \begin{tabular}[c]{@{}l@{}}Conv(144, 3, 3)\\MaxPool(2, 2)\end{tabular}
& \begin{tabular}[c]{@{}l@{}}Conv(256, 5, 5)\\MaxPool(2, 2)\end{tabular}
& \begin{tabular}[c]{@{}l@{}}FC(512)\\FC(L)\end{tabular}
& \begin{tabular}[c]{@{}l@{}}FC(1024)\\FC(1024)\\FC(1)\end{tabular}
\\
\bottomrule
\end{tabular}
\end{adjustbox}
\end{table}

\subsection{Training stage} \label{subsec:training}
To ensure a fair comparison with the original DANN algorithm, we set the same training configuration: Stochastic Gradient Descent with a learning rate of $0.01$, decay of $10^{-6}$, and momentum of $0.9$, as well as the same number of epochs ($300$).

For the iterative stage of iDANN, we set $\iEpochs$ to $25$. This value was determined empirically. We observed that it allowed the network weights to be tuned with the new knowledge without taking too long to perform a new iteration. In each training iteration, the greater improvement occurs in the first epochs, after which the accuracy of the label classifier is stabilized.

Concerning the size of the subset to select from the target set ($\iSize$), we decided to consider a percentage of the remaining samples rather than a fixed value. Initially, we set it to $5 \%$, and it was increased after each iteration by $150 \%$ ($\beta=1.5$) until all target domain samples are selected. This value was also obtained empirically, by observing better results and more stable training if few samples are added in the first iterations.

Different values for both the batch size $b$ and $\lambda$ are evaluated, as will be reported in the experimentation section. 



\section{Results}
\label{sec:experiments}

In this section we evaluate the proposed method using the datasets, topologies, and settings described in Section \ref {sec:setup}. We first study the different hyper-parameterization, as well as the two prototype selection policies proposed. Next we show the performance results obtained over the datasets and, finally, we compare with other state-of-the-art methods.

\subsection{Hyper-parameters evaluation}

In this section, we start by analyzing the influence of the batch size and the value of $\lambda$ on the performance of the method, as these hyper-parameters are those that affect the training stage the most. For this, we consider the batch sizes of $\{16, 32, 64, 128, 256, 512\}$ and $\lambda$ of  $\{10^{-1}, 10^{-2}, 10^{-3}, 10^{-4} \}$. This means that each result comes from a total of 336 experiments (14 combinations of dataset pairs $\times$ 6 batch values $\times$ 4 values of $\lambda$). The rest of hyper-parameters are set as indicated in Section \ref{subsec:training}, that is: $e = 300$ (as in the original DANN paper), $\iEpochs = 25$, and $\iSize = 5 \%$, which were empirically determined to favor stable training and obtain good results. In addition, we evaluate the results using only the prototype selection policy based on network's confidence, as next section will be devoted to comparing the two proposed policies with the best hyper-parameters found.

As we are dealing with an unsupervised method, we mainly focus on analyzing the trend when modifying these parameters. Table \ref{tab:gs-results} shows the results of this experiment, where each figure represents the average of the 14 possible combinations of source and target domain of the datasets considered and all the iterations performed by the iDANN algorithm.

The first thing to remark is that some of the hyper-parameter combinations evaluated in these experiments do not converge ($h = 32/64$, $\lambda = 10^1$ for traffic signs). This could be detected automatically, since the accuracy is abruptly reduced to a value approximately equal to a random guess, for both the training set and evaluation set and for both the source and the target domain. However, these results have been kept in order to observe the general trend of the method and how these parameters affect it.

It can also be observed that the best performance is achieved with $\lambda = 10^{-2}$ in the two types of corpora, while a batch size of $64$ and $32$ are better for the digits and traffic signs, respectively. On average, better results are reported with low $\lambda$ values and batch sizes between $32$ and $256$. When $\lambda$ is greater (e.g., $10^{-1}$), the training becomes highly unstable, especially if combined with small batch sizes.

\begin{table}[ht]
\caption{Influence of hyper-parameter setting on the performance (accuracy, in \%) of the iDANN algorithm. Figures report classification accuracy over the target set, averaging with the respective datasets and iterations of the algorithm.}
\label{tab:gs-results}
\centering
\scriptsize
\begin{tabular}{clcccclcccc}
\toprule
               & & \multicolumn{4}{c}{\textbf{Numbers}}           &   & \multicolumn{4}{c}{\textbf{Traffic signs}}       \\ \cline{3-6} \cline{8-11}
               & & \multicolumn{4}{c}{$\boldsymbol{\lambda}$}    &  & \multicolumn{4}{c}{$\boldsymbol{\lambda}$}     \\ \cline{3-6} \cline{8-11}
\textbf{Batch} & & $\boldsymbol{10^{-4}}$
                 & $\boldsymbol{10^{-3}}$
                 & $\boldsymbol{10^{-2}}$
                 & $\boldsymbol{10^{-1}}$

                 &  & $\boldsymbol{10^{-4}}$
                 & $\boldsymbol{10^{-3}}$
                 & $\boldsymbol{10^{-2}}$
                 & $\boldsymbol{10^{-1}}$
                     \\
                 \cline{1-1} \cline{3-6} \cline{8-11}
\textbf{16}    & & 58.74	      & 56.21	& 58.65	         & 47.54		         & & 89.67	& 88.65	& 90.08	         & 48.16 \\
\textbf{32}    & & 66.13	      & 65.82	& 61.67	         & 49.78		         & & 93.58	& 93.63	& \textbf{94.50} & 24.02	\\
\textbf{64}    & & 65.26	      & 66.41	& \textbf{66.82} & 62.54	 & & 91.27	& 91.16	& 91.67	         & 31.41	 \\
\textbf{128}   & & 64.23	      & 66.04	& 66.79	         & 52.89		         & & 88.56	& 89.36	& 89.60	         & 66.78	 \\
\textbf{256}   & & 64.55	      & 63.94	& 64.24	         & 59.36		         & & 87.34	& 87.73	& 88.67	         & 91.39	 \\
\textbf{512}   & & 62.55	      & 62.61	& 62.75	         & 50.67		         & & 84.34	& 84.45	& 84.09	         & 84.60 \\
\bottomrule
\end{tabular}
\end{table}

Next, we analyze the influence of these parameters with respect to the iteration of the iDANN algorithm. Table \ref{tab:it-increment} shows the average result obtained by grouping all combinations of datasets (numbers and traffic signs) and hyper-parameters considered. As in the previous analysis, better results are also observed for low $\lambda$ values and batch sizes between $32$ and $256$ (see column `Avg.'). In this case, it can also be seen that low $\lambda$ values are more appropriate in the first iterations, whereas greater $\lambda$ values are more appropriate in the last iterations. It might happen that a more stable way of proceeding (low $\lambda$) is preferred in the first iterations, even at the cost of being less aggressive in the domain adaptation. Therefore, we propose to start with a low $\lambda$ and increase its value gradually ($+10^{- 4}$ after each epoch).

Additionally, it is observed that each iteration of the algorithm leads to a better result than the previous one (except for $\lambda \ge 10^{-1}$), yielding the higher leap in the first iterations and reducing this difference towards the last iterations. Including all cases, the results improve by $5.19 \%$ between the first and the last iteration, on average. If we ignore those settings that do not converge, the average improvement obtained increases to $10.29 \%$.

\begin{table}[ht]
\caption{Influence of hyper-parameter setting on the performance of the iDANN algorithm with respect to number of iterations. Figures report classification accuracy over the target set, averaging with the respective datasets.}
\label{tab:it-increment}
\centering
\scriptsize
\begin{tabular}{cccccccccccc}
\toprule
                            &                & \multicolumn{9}{c}{\textbf{Iterations}}                                                                                         &      \\
\cline{3-11}
$\boldsymbol{\lambda}$      & \textbf{Batch} & \textbf{1} & \textbf{2} & \textbf{3} & \textbf{4} & \textbf{5} & \textbf{6} & \textbf{7} & \textbf{8} & \textbf{9} & \textbf{Avg.} \\ \hline
\multirow{6}{*}{$\boldsymbol{10^{-4}}$ }
            & 16   & 58.46	&	59.71	&	61.46	&	62.81	&	63.91	&	64.86	&	65.39	&	65.78	&	66.04	&	63.16 \\
            & 32   & 65.20	&	67.85	&	69.37	&	69.91	&	70.73	&	71.14	&	71.89	&	72.11	&	\textbf{72.24}	&	\textbf{70.05} \\
            & 64   & 63.88	&	67.11	&	68.33	&	68.75	&	69.35	&	70.30	&	70.71	&	71.13	&	71.22	&	68.97 \\
            & 128  & 62.67	&	65.52	&	67.09	&	67.68	&	68.19	&	68.84	&	69.46	&	69.88	&	70.06	&	67.71 \\
            & 256  & 62.82	&	65.20	&	66.92	&	67.83	&	68.65	&	68.84	&	69.57	&	70.09	&	70.34	&	67.81 \\
            & 512  & 61.51	&	63.28	&	64.32	&	65.45	&	65.97	&	66.93	&	67.60	&	67.87	&	68.03	&	65.66 \\
\hline
\multirow{6}{*}{$\boldsymbol{10^{-3}}$ }
            & 16   & 56.65	&	57.70	&	59.65	&	60.72	&	61.43	&	62.25	&	62.75	&	63.15	&	63.31	&	60.85 \\
            & 32   & 63.95	&	67.42	&	68.68	&	69.93	&	70.59	&	71.26	&	71.79	&	72.19	&	72.33	&	69.79 \\
            & 64   & 64.66	&	67.39	&	68.78	&	69.89	&	70.41	&	71.32	&	72.06	&	72.44	&	\textbf{72.57}	&	\textbf{69.95} \\
            & 128  & 63.75	&	66.59	&	68.61	&	69.07	&	69.93	&	70.82	&	71.44	&	72.00	&	72.14	&	69.37 \\
            & 256  & 62.38	&	65.08	&	66.67	&	67.44	&	67.69	&	68.52	&	69.10	&	69.49	&	69.71	&	67.34 \\
            & 512  & 61.36	&	63.46	&	64.72	&	65.48	&	66.25	&	66.96	&	67.42	&	67.84	&	68.08	&	65.73 \\
\hline
\multirow{6}{*}{$\boldsymbol{10^{-2}}$ }
            & 16   & 56.73	&	61.57	&	63.37	&	65.13	&	65.81	&	66.31	&	62.94	&	63.13	&	63.26	&	63.14 \\
            & 32   & 62.07	&	64.75	&	66.01	&	66.68	&	67.46	&	67.78	&	68.23	&	68.47	&	68.80	&	66.69 \\
            & 64   & 64.78	&	67.77	&	69.44	&	70.20	&	71.21	&	71.92	&	72.35	&	72.74	&	\textbf{72.95}	&	\textbf{70.37} \\
            & 128  & 64.49	&	67.27	&	69.02	&	69.75	&	70.71	&	71.58	&	72.00	&	72.72	&	72.87	&	70.05 \\
            & 256  & 63.16	&	65.55	&	66.75	&	67.49	&	68.32	&	68.76	&	69.51	&	69.81	&	70.21	&	67.73 \\
            & 512  & 61.61	&	63.49	&	64.82	&	65.57	&	66.10	&	66.81	&	67.52	&	68.01	&	68.28	&	65.80 \\
\hline
\multirow{6}{*}{$\boldsymbol{10^{-1}}$ }
            & 16   & 46.48	&	50.71	&	52.12	&	53.35	&	53.78	&	42.41	&	42.98	&	43.35	&	43.52	&	47.63 \\
            & 32   & 50.09	&	53.07	&	47.66	&	42.61	&	43.12	&	44.06	&	44.39	&	44.91	&	44.96	&	46.10 \\
            & 64   & 61.64	&	64.02	&	64.24	&	54.01	&	54.48	&	55.44	&	55.96	&	56.47	&	56.59	&	58.09 \\
            & 128  & 55.72	&	57.10	&	57.16	&	57.01	&	52.95	&	53.15	&	53.63	&	53.50	&	53.69	&	54.88 \\
            & 256  & 60.13	&	62.26	&	62.60	&	63.53	&	64.25	&	64.93	&	65.57	&	66.04	&	\textbf{66.14}	&	\textbf{63.94} \\
            & 512  & 53.54	&	54.39	&	54.63	&	55.01	&	55.69	&	56.04	&	56.57	&	56.84	&	56.95	&	55.52 \\
\hline
\multicolumn{2}{c}{\textbf{Average}}
                   & 60.32	&	62.84	&	63.85	&	63.97	&	64.46	&	64.63	&	65.03	&	65.42	&	\textbf{65.59}	&	-- \\
\bottomrule
\end{tabular}
\end{table}

\subsection{Model analysis}
\begin{figure*}[ht]
  \centering
  \includegraphics[width=.6\linewidth]{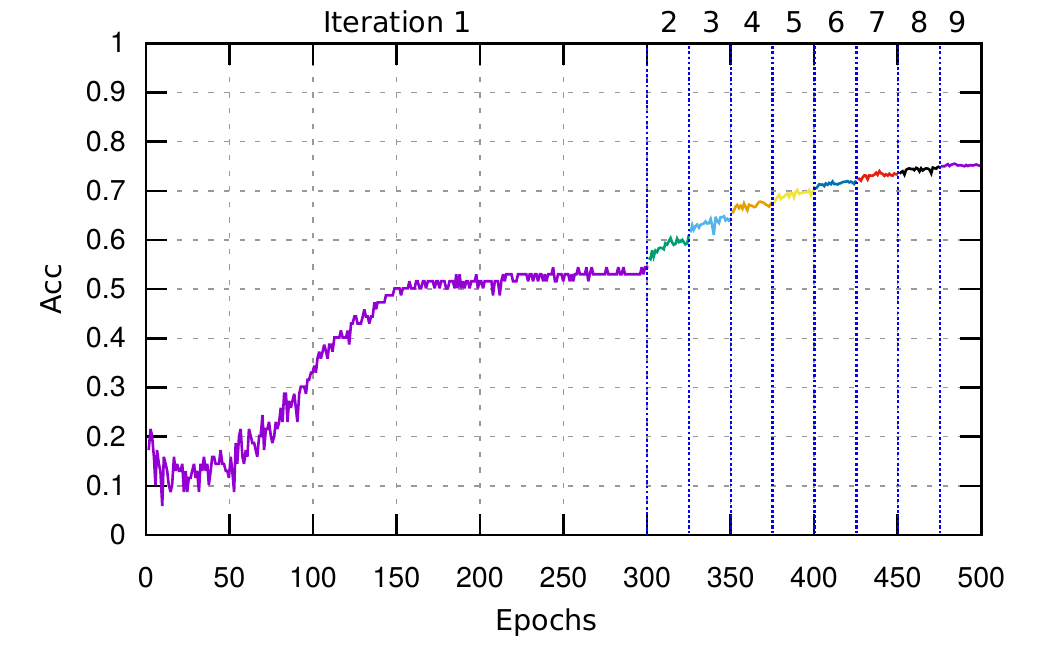}
  \caption{Accuracy curve with respect to training epochs and iterations of the incremental approach.}
  \label{fig:plot_training}
\end{figure*}

We now evaluate the effect of the incremental training process on the domain adaptation approach. Figure \ref{fig:plot_training} shows the evolution of the accuracy obtained over the target test set during the training process for the case Syn Numbers $\rightarrow$ MNIST-M combination of datasets, with a batch size of 64 and $\lambda = 10^{-2}$. The training epochs are represented with the horizontal axis, while the iterations (i.e., when new training samples are added) are highlighted with blue lines and marked above. It can be observed that in the first iteration (spanning 300 epochs), the accuracy slowly improves until around 150 epochs, after which becomes stable. In the subsequent iterations, the accuracy further improves, especially during iterations 2, 3 and 4. Then, the performance increase is gradually reduced until it is hardly noticeable.

\begin{figure*}[ht]
  \centering
  \includegraphics[width=1\linewidth]{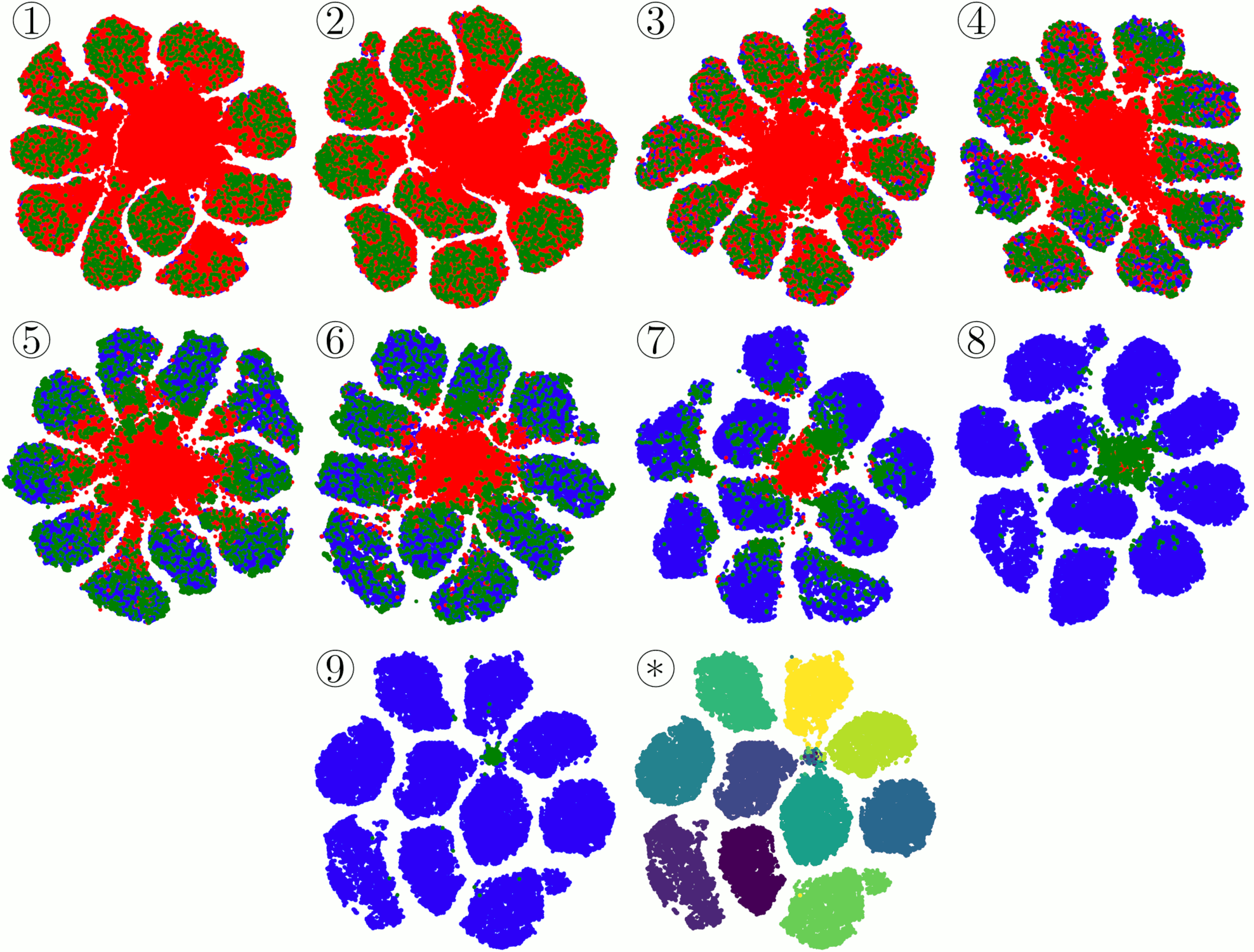}
  \caption{t-SNE representation of the feature space of the neural network (from the last hidden layer) with respect to the iteration of the approach. The red color represents the target domain, the blue color represents the source domain, and the green color represents the set $\hat{B}_{\iSize}$. Last representation (*) depicts each sample according to its actual category by using different colors.}
  \label{fig:tsne}
\end{figure*}

To provide further analysis, we also examine the representation space learned by the network in each of these iterations, using the same combination of datasets and training parameters. We use the t-Distributed Stochastic Neighbor Embedding (t-SNE)~\cite{vanDerMaaten2008tsne} projection to visualize the samples according to their representation by the last hidden layer of the label predictor. Figure \ref{fig:tsne} shows a visualization of the features learned after each of the iterations, where the red color represents the target domain, the blue color represents the source domain, and the green color represents the set $ \hat{B}_{\iSize}$ (selected samples) using the confidence policy. This representation reveals $10$ well-defined clusters---the 10 possible classes of the datasets considered for this analysis---around an additional central cluster. This central cluster groups the samples of the target domain (red color) whose representation does not correspond to any of the existing classes yet. This cluster would therefore correspond to target samples whose representation has not been correctly mapped onto any of the source domain classes. Iteratively, the method is selecting samples (green points) of the target domain and moving them to the source domain. In the first iterations---until the 6th one, approximately---the method selects only samples that are well located in one of the source domain clusters (that is, those samples for which the network is more confident). Due to this process, the size of the central cluster is reduced. It is important to emphasize that this cluster becomes smaller although no samples out of it are selected, which indicates that the network is learning to better map those samples because of the selected samples of previous iterations. Towards the last iterations, the method begins to select the most complex samples that are still in this additional cluster. In Fig.~\ref{fig:tsne}(*) (which is the same as the Fig. \ref{fig:tsne}(9) but highlighting each class with a different color), the additional cluster of target samples still appears without being mapped, yet with a very small size. This cluster contains almost all the classification errors, having mapped only some isolated prototypes to the actual class clusters incorrectly.

\subsection{kNN policy}
We compare in this section the two policies proposed for selecting the set of target prototypes $\hat{B}_{\iSize}$ to be added to the source domain. To this end, we evaluate whether the label assigned to each of these prototypes is correct. In this case, we make use of the ground-truth of the target domain just for the sake of analysis.

\begin{figure*}[ht]
  \centering
  \includegraphics[width=.7\linewidth]{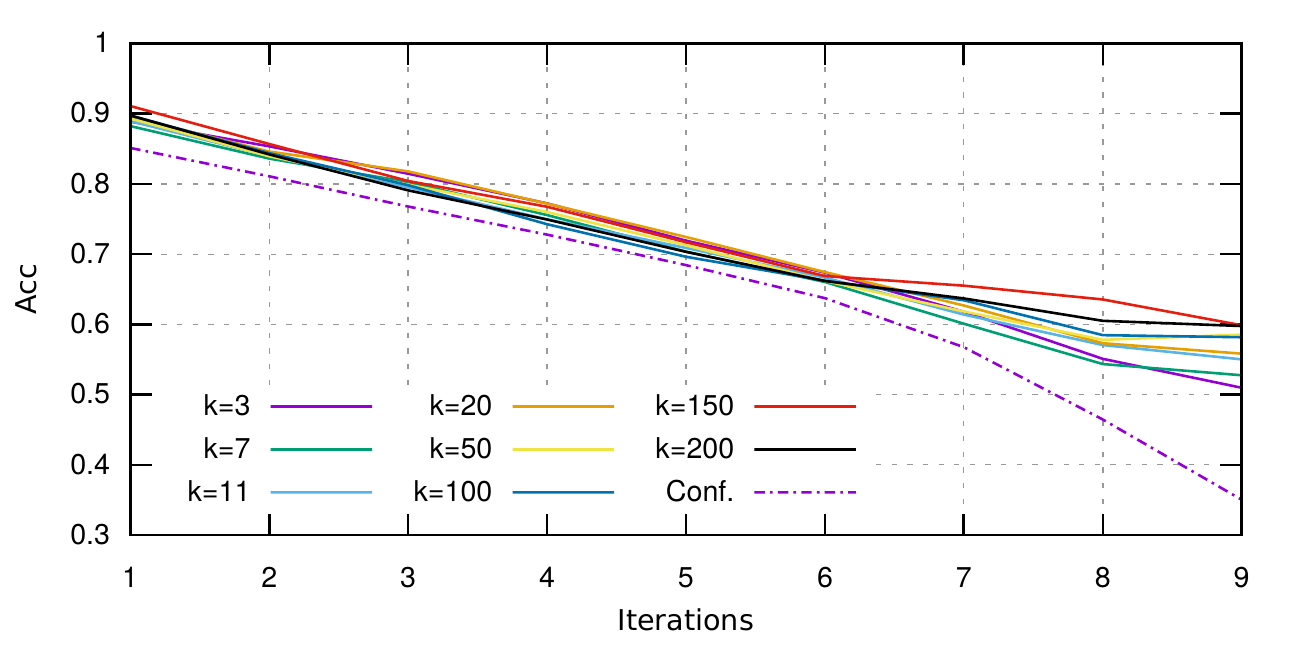}
  \caption{
  Accuracy of the labels assigned to the selected prototypes set ($\hat{B}_{\iSize}$) from the target domain according to the selection policy.
  }
  \label{fig:result_knn}
\end{figure*}

We show in Fig. \ref{fig:result_knn} a dotted line with the performance of the confidence policy, which may serve as a baseline here, and eight results for the kNN policy with varying $k$ values. As in the previous experiments, the reported figures are obtained for all combinations of datasets and hyper-parameters considered.

It is observed that, as the number of iterations of the algorithm increases, the accuracy of the additional labels assigned to the selected prototypes decreases. However, the kNN policy generally obtains better results from the first iteration, obtaining on average (for all iterations) an improvement of 6.36~\% with respect to confidence policy. This improvement is significantly greater in the last iterations, obtaining an increase up to 24.85~\% between the result of the confidence policy and the best result obtained with kNN policy.

The role of the parameter $k$ is also illustrated in Fig. \ref{fig:result_knn}, where better results are attained as $k$ is increased. It is shown that the impact of this parameter is more noticeable in the last iterations, where a difference of up to 8.91~\% is obtained between $k = 3$ and $k = 150$.

Because the kNN selection policy worked better, this policy was used in all following experiments.

\subsection{Accuracy on target}


In this section, we evaluate the final result obtained through the proposed iDANN method with the best combination of hyper-parameters previously obtained for each of the dataset pairs. We will compare this result with that obtained by the original DANN method in order to check the goodness of the incremental approach.

Table \ref{tab:results} reports the results of the experiment, where rows indicate the dataset pairs (source and target) and columns represent the DA method. Concerning iDANN, we report two results: the accuracy of the labels assigned during the iterative process itself (1), as well as the accuracy using the CNN trained from scratch using only the target samples (once all the target samples have been assigned a label). In addition to DANN and iDANN methods, we have also added the results obtained with the neural networks trained just with the source set (`CNN Src.'), as well as the results obtained with the neural neural networks directly trained with the target set (`CNN Tgt.'). The former serves as baseline, to better assess the impact of the domain-adaptation mechanisms, while the second represents the upper bound of accuracy.

The first thing to remark is that the worst results obtained by the baseline (`CNN Src.') come from the combinations of single-digit datasets (MNIST, MNIST-M) as source and complex digit datasets (SVHN, Syn Numbers) as target. Furthermore, the best results from the baseline are reported for combinations where the source and target are similar (MNIST-M $\rightarrow$ MNIST,  SVHN $\rightarrow$ Syn Numbers).

The original DANN method outperforms the results obtained by using the baseline network (`CNN Src.') by 10.7~\%, on average, obtaining the most significant improvement for the combinations of Syn Numbers $\rightarrow$ MNIST (improving by 29.31~\%). It is also noticeable the impact of DANN when the dataset pair consists of similar tasks with the most complex one as target, such as MNIST $\rightarrow$ MNIST-M---improvement of 23~\%---or Syn Signs $\rightarrow$ GTSRB---improvement of 15.49~\%. These results for DANN have been obtained using our own implementation, following the details given in the original paper. We observed that the accuracy matches approximately that reported by the authors (for the 4 combinations they considered), and so we assume that our implementation is correct. We can therefore faithfully report the performance in all source-target combinations of our experiments.


Concerning the labels assigned during the proposed incremental approach iDANN (${D_T}\text{Acc.}^{(1)}$), the first thing to note is its improvement with respect to the underlying DANN method, which is around 16~\%, on average. In the best case, this improvement reaches values around 33~\%, 35~\% and 36~\% for the Syn Numbers $\rightarrow$ MNIST-M, MNIST-M $\rightarrow$ Syn Numbers, and MNIST $\rightarrow$ Syn Numbers pairs, respectively. This confirms the goodness of our strategy, which uses the same domain adaptation method in a novel way.

Finally, if the CNN is trained from scratch with the target labels that have been automatically assigned by the iDANN ($D_T \text{Acc.}^{(2)}$), it can further improve the results up to 1.64~\%, on average, and up to 5.5~\% in the best case (MNIST-M $\rightarrow$ Syn Numbers). It should be noted that in some specific combinations, this approach slightly outperforms the CNN trained with the correct target labels (for example, MNIST-M $\rightarrow$ MNIST or Syn Numbers $\rightarrow$ SVHN). It might happen that the incorrectly assigned labels of the iDANN process act as a regularizer that alleviates some overfitting.

\begin{table}[ht]
\caption{Accuracy (\%) over the the target dataset for different strategies: `CNN Src.' indicates a neural network trained only with source samples; `DANN' denotes the original DANN strategy; `iDANN' yields two results from the incremental strategy: $D_T\text{Acc.}^{(1)}$ refers to the accuracy of the labels assigned to the target samples during the iterative process, while $D_T\text{Acc.}^{(2)}$ refers to the classification after training a new CNN from scratch using the labels assigned to the target samples; and `CNN Tgt.' denotes a CNN trained using the ground truth of the target samples.}
\label{tab:results}
\centering
\begin{adjustbox}{width=\textwidth, keepaspectratio}
\begin{tabular}{c|c|c|c|cc|c}
\toprule
\multirow{2}{*}{\textbf{Source}}
    & \multirow{2}{*}{\textbf{Target}}
    & \textbf{CNN Src.}
    & \textbf{DANN}
    & \multicolumn{2}{c|}{\textbf{iDANN}}
    & \textbf{CNN Tgt.}
    \\
\cline{3-7}
& & $\boldsymbol{D_T}$ \textbf{Acc.}
  & $\boldsymbol{D_T}$ \textbf{Acc.}
  & $\boldsymbol{D_T} \textbf{Acc.}^{(1)}$ & $\boldsymbol{D_T} \textbf{Acc.}^{(2)}$
  & $\boldsymbol{D_T}$ \textbf{Acc.}
  \\
\hline
\multirow{3}{*}{MNIST}
    & MNIST-M           & 55.71
                        & 78.70
                        & 96.09
                        & 96.67
                        & 97.34
                        \\
    & SVHN              & 16.26
                        & 31.32
                        & 35.83
                        & 36.49
                        & 90.93
                        \\
    & Syn Numbers       & 32.14
                        & 44.66
                        & 80.79
                        & 84.82
                        & 99.34
                        \\
\hline
\multirow{3}{*}{MNIST-M}
    & MNIST             & 97.95
                        & 98.65
                        & 99.04
                        & 99.59
                        & 98.94
                        \\
    & SVHN              & 32.91
                        & 41.41
                        & 61.87
                        & 61.89
                        & 90.93
                        \\
    & Syn Numbers       & 46.34
                        & 54.02
                        & 89.49
                        & 94.99
                        & 99.34
                        \\
\hline
\multirow{3}{*}{SVHN}
    & MNIST             & 59.04
                        & 67.08
                        & 82.72
                        & 84.50
                        & 98.94
                        \\
    & MNIST-M           & 43.49
                        & 47.42
                        & 66.40
                        & 67.62
                        & 97.34
                        \\
    & Syn Numbers       & 88.42
                        & 89.56
                        & 96.43
                        & 98.10
                        & 99.34
                        \\
\hline
\multirow{3}{*}{Syn Numbers}
    & MNIST             & 60.04
                        & 89.35
                        & 98.13
                        & 99.35
                        & 98.94
                        \\
    & MNIST-M           & 41.84
                        & 54.38
                        & 87.10
                        & 90.26
                        & 97.34
                        \\
    & SVHN              & 85.16
                        & 87.24
                        & 91.42
                        & 91.95
                        & 90.93
                        \\
\hline
GTSRB     &	Syn signs   & 76.39
                        & 86.22
                        & 98.28
                        & 98.57
                        & 99.74
                        \\
\hline
Syn signs & GTSRB       & 69.79
                        & 85.28
                        & 96.31
                        & 98.00
                        & 97.89
                        \\
\hline
\hline
\multicolumn{2}{c|}{Average}
                        & 57.53
                        & 68.23
                        & 84.28
                        & 85.91
                        & 96.95
                        \\
\bottomrule
\end{tabular}
\end{adjustbox}
\end{table}

\subsection{Comparison with the state of the art
\label{sec:comparison}}

To conclude the results section, we present below a comparison with other domain adaptation strategies from the state of the art. In these works, not all possible combinations of source-target pairs are considered but a few combinations of them. We show in Table \ref{tab:comparison} the results reported in the literature\footnote{Unlike the results of the previous section, the DANN values of Table \ref{tab:comparison} are those reported in the original paper \cite{Ganin2016DANN}.}, along with the results obtained by our proposal (iDANN). A brief description of the competing methods was provided in Section \ref{sec:background}. Readers are referred to the corresponding references for details.


These results reveal that our method yields the best performance in 5 out of 7 source-target pairs. The performance of iDANN is especially remarkable in the case of MNIST $\rightarrow$ Syn Num, where the improvement reaches around 30~\% compared to the literature.
For the case in which our proposal does not attain the best result, we observe a dissimilar performance: it is still very competitive for the MNIST $\rightarrow$ SVHN pair, whereas it is outperformed for the SVHN $\rightarrow$ MNIST pair. When all the results are good, the improvement is relative, but when there is enough margin, the improvement is quite remarkable (as in the case of MNIST $\rightarrow$ Syn Num).

\begin{table}[ht]
\caption{Comparison of accuracy (\%) between state-of-the-art DA approaches and iDANN. The first two rows denote the source and target dataset, respectively. The best result for each combination (column) is highlighted in bold typeface. Empty cells indicate that the result is not reported in the literature.}
\label{tab:comparison}
\centering
\begin{adjustbox}{width=\textwidth, keepaspectratio}
\begin{tabular}{lccclcclcc}
\toprule
                              & \multicolumn{3}{c}{\textbf{MNIST}}                  & & \multicolumn{2}{c}{\textbf{SVHN}} & & \multicolumn{2}{c}{\textbf{Syn Num}} \\ \cline{2-4} \cline{6-7}  \cline{9-10}
\textbf{Methods}              & \textbf{MNIST-M} & \textbf{SVHN} & \textbf{Syn Num} & & \textbf{MNIST} & \textbf{Syn Num} & & \textbf{MNIST}    & \textbf{SVHN}    \\
\midrule
SA \cite{Fernando2013SA}      & 56.9             & --            & --               & & 59.32          & --               & & --                & 86.44            \\
DRCN \cite{Ghifary2016}       & --               & \textbf{40.05}& --               & & 81.97          & --               & & --                & --               \\
DSN \cite{Bousmalis2016}      & 83.2             & --            & --               & & 82.7           & --               & & --                & 91.2             \\
DANN \cite{Ganin2016DANN}     & 76.66            & 12.4          & 22.9             & & 73.85          & 96.9             & & 87.6              & 91.09            \\
D-CORAL \cite{Baochen2016CORAL} & --               & 35            & 55.8             & & 76.3           & 95.5             & & 89.9              & 78.8             \\
ADDA \cite{Tzeng2017}         & --               & --            & --               & & 76             & --               & & --                & --               \\
ADA \cite{Haeusser2017}       & --               & 12.9          & 34.8             & & 96.3           & 95.5             & & 97.1              & 88.1             \\
VADA \cite{shu2018a}          & --               & 18.6          & 45.9             & & 92.9           & 96.8             & & 96.2              & 85.3             \\
DeepJDOT \cite{Damodaran2018} & 92.4             & --            & --               & & 96.7           & --               & & --                & --               \\
DTA \cite{Lee2019}            & --               & --            & --               & & \textbf{99.4}  & --               & & --                & --               \\
\midrule
iDANN                         & \textbf{96.67}   & 36.49         & \textbf{84.82}   & & 84.50          & \textbf{98.10}   & & \textbf{99.35}    & \textbf{91.95}   \\
\bottomrule
\end{tabular}
\end{adjustbox}
\end{table}

Furthermore, it should be noted that many of the compared methods propose specific CNN architecture for each combination of datasets and/or focus on optimizing the result for a particular combination, such as DTA or DeepJDOT. In our case, we utilized the topologies proposed in the original DANN paper, so it could be assumed that if we pursue a specific architecture adapted to each of the source-target pairs, our results will probably improve.

\section{Conclusions and Future Work}
\label{sec:conclusions}

This paper proposes an incremental strategy to the problem of domain adaptation with artificial neural networks. Our approach is built upon an existing domain adaptation approach, combined with a heuristic that, in each iteration, decides which prototypes of the target set can be added to the training set by considering the label provided by the neural network. To this end, two selection policies have been proposed: one directly based on the confidence given by the network to the prediction and another based on geometric properties of the learned feature space. We observed that the latter reported a better performance, especially in the last iterations of the algorithm. In addition, we consider a final stage in which the labeled target set is used to train a new neural network with label smoothing.

Our experiments were performed on various corpora and using several configurations of the neural network. From the results, we conclude that the incremental approach outperforms the underlying DANN model, as well as other state-of-the-art methods. It is interesting to note that, in some cases, the iDANN approach improves the result obtained with the CNN trained directly with the ground-truth data of the target set, which could indicate that the incremental process also serves as a regularizer that leads to greater robustness. Furthermore, unlike the classic DANN, our approach improves results when domains are similar and helps keeping the accuracy for the source domain. We also observed a greater training stability and less dependence on the hyper-parameters set.

As future work, a primary objective would be to establish a well-principled stop criterion that allows us to detect when the prediction over the target samples is not reliable. In addition, we want to extend the experiments to other types of input types (such as sequences), as well as to study the behavior of the incremental strategy when the underlying DA method is different---given that there currently exist several architectures for this challenge. Note that our incremental approach is independent of the underlying DA model considered, and so it could be adopted as a generic strategy that might improve to the same extent as the underlying DA algorithm improves. Other avenues to further explore this proposal is to evaluate more neural network architectures, as well as adding data augmentation to the learning process.


\bibliographystyle{model5-names} 

\bibliographystyle{elsarticle-num}

\bibliography{mybibfile}

\end{document}